\documentclass{article}
\usepackage{spconf,amsmath,graphicx}
\usepackage{multirow}

\title{HIERARCHICAL CROSS NETWORK FOR PERSON RE-IDENTIFICATION}

\name{Huan-Cheng Hsu$^1$, Ching-Hang Chen$^2$, Hsiao-Rong Tyan$^3$, Hong-Yuan Mark Liao$^{1,3}$}
\address{$^1$Institute of Information Science, Academia Sinica, Taiwan\\
	$^2$Amazon Inc. Cupertino, CA, USA\\
	$^3$Department of Information and Computer Engineering, Chung Yuan University Taiwan}

\begin{document}
%\ninept
%
\maketitle
\begin{abstract}
Person re-identification (person re-ID) aims at matching target person(s) grabbed from different and non-overlapping camera views. It plays an important role for public safety and has application in various tasks such as, human retrieval, human tracking, and activity analysis. In this paper, we propose a new network architecture called Hierarchical Cross Network (HCN) to perform person re-ID. In addition to the backbone model of a conventional CNN, HCN is equipped with two additional maps called hierarchical cross feature maps. The maps of an HCN are formed by merging layers with different resolutions and semantic levels. With the hierarchical cross feature maps, an HCN can effectively uncover additional semantic features which could not be discovered by a conventional CNN. Although the proposed HCN can discover features with higher semantics, its representation power is still limited. To derive more general representations, we augment the data during the training process by combining multiple datasets. Experiment results show that the proposed method outperformed several state-of-the-art methods.

\end{abstract}
\begin{keywords}
Person re-identification, Deep learning, Convolutional Neural Networks
\end{keywords}

\section{Introduction}
\label{sec:intro}

Person re-ID aims at matching target person(s) grabbed from different and non-overlapping camera views. It plays an important role for public safety and has application in various tasks such as, human retrieval, human tracking, and activity analysis. Person re-ID is a very challenging task since it has to deal with many difficult problems, for example, low resolution of video camcorder, occlusion among objects, large intra-class scatter, etc. To tackle the above mentioned issues, learning robust and invariant features from a target image is important.

Recently, since the rise of deep learning, the performance on person re-ID has also dramatically been improved and many deep learning-based methods \cite{li2014deepreid,zheng2016person,chen2016similarity,liu2017end} have been proposed. Although the structure of a conventional convolutional neural network(CNN) is able to powerfully learn general and semantic features from a target scene, there are remain abundant un-use information distributed over layers of the networks.

%%%%%%%%%%%%%%%%%%%%%%%%%%%%fig.1
\begin{figure}[tbp]

\begin{minipage}[b]{1.0\linewidth}
  \centering
  \centerline{\includegraphics[width=7.0cm]{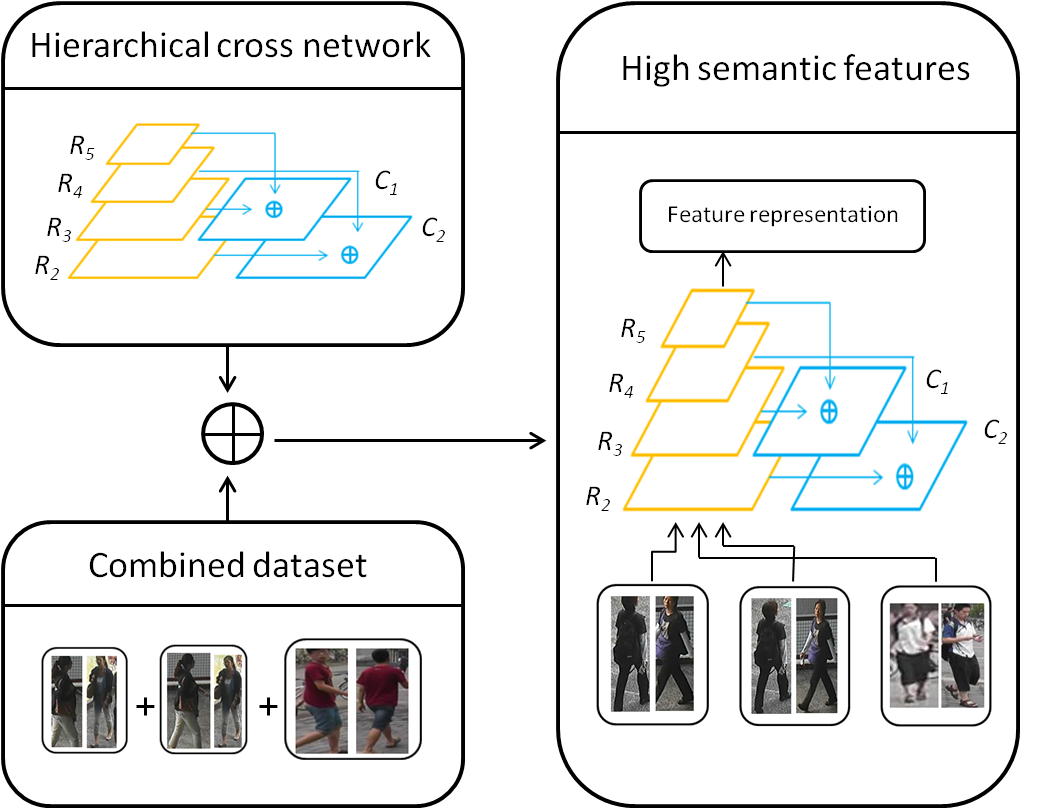}}
  \vspace{0.1cm}
  \centerline\medskip
\end{minipage}
\caption{Overview of our framework for person re-ID. It consists two parts: a systematic model and a strategy on training which are the hierarchical cross network and the combined dataset respectively. In  the first part, $\mathit{\{R_2,R_3,R_4,R_5\}}$ are feature maps of backbone and $\mathit{\{C_1, C_2\}}$ are hierarchical cross feature maps. In the second part, three image pairs denote datasets we used in experiments.}
\label{overview}

\end{figure}
%%%%%%%%%%%%%%%%%%%%%%%%%%%

To uncover the information embedded in layers of a CNN in a more efficient manner, several methods have been proposed~\cite{hariharan2015hypercolumns,long2015fully,lin2017feature}. In~\cite{hariharan2015hypercolumns}, Hariharan \emph{et al}. proposed the hypercolumn concept which asserts the concatenation of features from feature maps and makes it a long vector for matching. Long \emph{et al}.~\cite{long2015fully} proposed the use of a DAG net to fuse the information coming from the fine layers and the coarse layers, respectively, and then obtain both local and global structures for further use. Lin \emph{et al}.~\cite{lin2017feature} proposed a feature pyramid network which exploits the inherent multi-scale information in network to enhance the object representation power. To better discover un-use information embedded in a CNN, we proposed a new network architecture called hierarchical cross network (HCN) to achieve the goal. In addition to the backbone model of a conventional CNN, HCN is equipped with two additional maps called hierarchical cross feature maps.

As shown in Figure \ref{overview}, the maps of an HCN are formed by merging layers with different resolutions and semantic levels. With the hierarchical cross feature maps, an HCN can effectively uncover additional semantic features which could not be discovered by a conventional CNN. 

Although the proposed HCN can discover features with higher semantics, its representation power is still limited. To derive more general representation, we augment the data during the training process by combining multiple datasets. Unlike many applications which deal with either classification or retrieval only, the task of person re-ID requires to handle both. Therefore, we divide the dataset into two non-overlapping subsets.
We make the number of identities trained in the training stage a variable since this number will not be limited by the number of unseen identities during retrieval. Therefore, augmenting the dataset by combining many related datasets is feasible. With this combined dataset concept realized, the time used for training data can be significantly reduced and the feature defined can be applied to cross-domain applications.

To summarize, our contributions include: (a) learning higher semantic and general features via a systematic framework and (b) getting competitive and efficient results comparing with state-of-the-art methods. Details are shown in experiments.

%%%%%%%%%%%%%%%%%%%%%%%%%%%%flowchart
\begin{figure*}[tbp]

\begin{minipage}[b]{1.0\linewidth}
  \centering
  \centerline{\includegraphics[width=17.0cm]{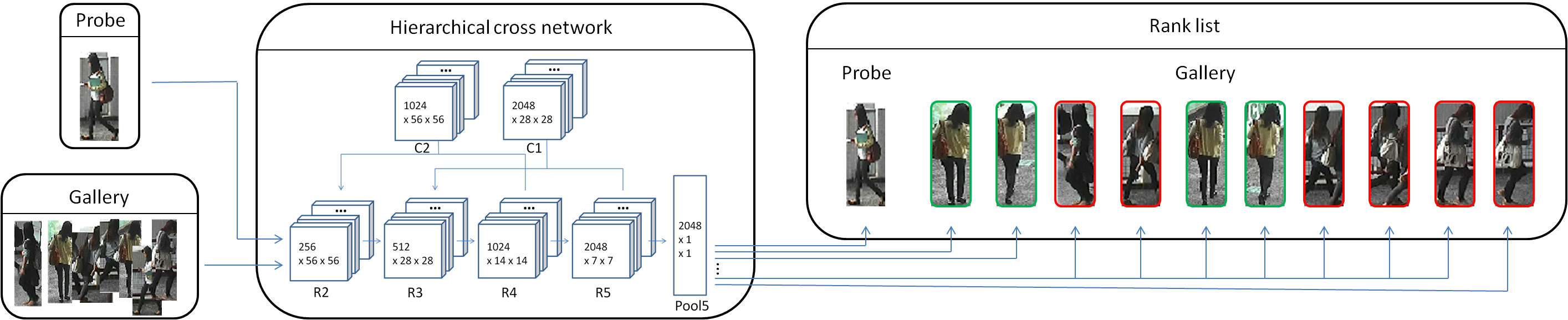}}
  \vspace{0.1cm}
  \centerline\medskip
\end{minipage}
\caption{The task of person re-ID is matching the target person(probe) from different and non-overlapping camera views(gallery). However, robust feature representations are needed due to many intensive appearance
changes such as lighting, pose, and viewpoint in person re-ID. We propose HCN to address this challenge. The left part are the probe and the gallery set which can be viewed as inputs to HCN. HCN provides high semantic  features for the probe and the gallery set shown in the middle part. In the right part, both probe and gallery sets are converted into high semantic features by HCN, which is expected to benefit performance in ranking.}
\label{flowchart}
\end{figure*}
%%%%%%%%%%%%%%%%%%%%%%%%%%%%%

\section{related works}
\label{sec:format}
\subsection{Person re-ID using hand-crafted features}
In the very begining, people extract hand-crafted features to perform person re-ID. In~\cite{farenzena2010person}, Farenzena \emph{et al}. proposed the use of weighted color histogram, most stable color regions, and recurrent high-structured patches to extract person from an image. Gheissari \emph{et al}.~\cite{gheissari2006person} proposed the use of HS histogram and edgel histogram within local regions to encode dominant local boundary orientation and RGB ratios for person re-ID. Besides the use of low-level features, middle-level features such as attribute-based features is also a possibility to use. In~\cite{shi2015transferring}, Shi \emph{et al}. proposed to learn attributes like color, texture, and category labels from fashion photography datasets.

\subsection{Person re-ID using deep-learning-based}
In contrast to the use of hand-crafted features for person re-ID, deep learning-based approaches adopt a completely different mechanism to handle the person re-ID problem.  Li \emph{et al}.~\cite{li2014deepreid} proposed the use of filter pairing neural network to handle misalignment, photometric and geometric transforms, occlusions, and cluttered background. In~\cite{liu2017end}, Liu \emph{et al}. tried to combine both soft attention and siamese model to ensure the focus is on the important parts of an input image pair. Cheng \emph{et al}.~\cite{chen2016similarity} proposed a triplet function which examines three input images at the same time, a positive pair to learn similar relationship and a negative image to learn dissimilar relationship.

\section{method}
\label{sec:pagestyle}
In this section, we shall describe the details of the proposed method. The problem statement will be given first, followed by the introduction of hierarchical cross network. Then, how a combined dataset is formed will be introducted in Section 3.3. In section 3.4, a re-ranking strategy will be detailed.

\subsection{Problem statement}
Given a probe image (contains a person), the goal of person re-ID is to match his/her counterpart in the gallery set. A person re-ID process requires to handle both classification and retrieval simultaneously. Therefore, our system design must consider this requirement. In the training stage, suppose that a dataset has $\mathit{M}$ identities with $\mathit{n}$ images ($\mathit{M}$ \textless $\mathit{n}$). Let $\mathit{D_i}$ = $\mathit{\{x_i, d_i\}}$ be the training set, where $\mathit{x_i}$ denotes the $\mathit{i}$-th image and $\mathit{d_i}$ denotes the identity covered in $\mathit{x_i}$. Given an unknown probe image $\mathit{x}$, we compute its feature representation $\mathit{f}$ and compare $\mathit{f}$ with the gallery set. Let the output of the comparison results be $\mathit{\textbf{z} = [z_1, z_2, ...,z_M]\in{R^M}}$. If we adopt Residual Network 50 (ResNet-50), the output $\mathit{\textbf{z}}$ will be FC1000 layer and its dimension is not fixed. Usually, the dimension of $\mathit{\textbf{z}}$ is equal to the number of identities covered in the training set. If the pool5 layer is adopted as feature representations, then its dimension is fixed.
The cross entropy loss of identity classification can be formulated as $L_{id}(f,d)=\sum_{m=1}^{M}-\log(p(m))q(m) $, where the predicted probability of each $\mathit{id}$ is calculated as: $\mathit{p(m|x)=\frac{\exp(z_m)}{\sum_{i=1}^{m}\exp(z_i)}}$, where $\mathit{q(y) = 1}$ if $\mathit{y}$ is the ground truth and $\mathit{q(m) = 0}$ for $\mathit{m \neq y}$.

In the testing stage, given a probe $\mathit{p}$ and a gallery set with $\mathit{K}$ images $\mathit{\{g_i\}_{i=1}^K}$, identity is determined by\begin{equation}\mathit{d_p=\operatornamewithlimits{argmax}_{i\in\{1,2,...,K\}}sim(q,g_i)}\end{equation}, where $\mathit{d_p}$ is the identity of probe $\mathit{q}$ and the similarity between $\mathit{p}$ and $\mathit{g_i}$ can be measured by Mahalanobis distance $\mathit{sim(p, g_i)=(f_p-f_{gi})\mathbf{M}^\top(f_p-f_{gi})}$ where $\mathbf{M}$ is a positive semidefinite matrix, $\mathit{f_p}$ and $\mathit{f_{gi}}$ represent computed descriptors of probe $\mathit{p}$ and gallery $\mathit{g_i}$, respectively.

\subsection{Hierarchical cross network}
In this work, we use ResNet-50~\cite{he2016deep} as the backbone of HCN. Although ResNet-50 has many residual blocks, we use the outputs of each stage's last residual block $\mathit{\{R_2,R_3,R_4,R_5\}}$ since these outputs have the strongest feature in each stage and we do not use conv1 since its semantically weak features do not bring better results. 

There are a number of existing methods~\cite{hariharan2015hypercolumns,long2015fully,lin2017feature} that also exploit multiple layers to increase the semantic level of representations. We follow the concept proposed in Lin \emph{et al}.~\cite{lin2017feature} to construct the feature maps. These maps are obtained by merging upsampled, semantically stronger feature maps and the corresponding bottom-up feature maps through lateral connections. However, we do not consider to merge layers iterated from the coarsest to the finest since adjacent layers are strongly correlated. To obtain high semantic feature maps, merging these correlated adjacent layers iteratively is not the best choice since the merged maps have low diversify on semantic. In practice, we apply skip connections to merge layers which contain additionally a layer between them. Specifically, the hierarchical cross feature maps $\mathit{\{C_1,C_2\}}$ do merge feature maps of backbone $\mathit{\{R_3,R_5\},\{R_2,R_4\}}$ by element-wise addition respectively.

Merging layers not only consider the height and the width of layers but also the depth of channel. In our work, we increase the capacity of HCN by extending channel dimensions of bottom-up maps to fit the channel dimensions of its corresponding top-down feature maps. Specifically, the depth of $\mathit{C_1}$ and $\mathit{C_2}$ are different, which are 2048 and 1024 respectively. To reduce the overfitting of HCN, dropout layers are added before the predictions of hierarchical cross feature maps $\mathit{C_1}$, $\mathit{C_2}$ and the layer of backbone $\mathit{R_5}$. As the setting mentioned above, HCN finally has three outputs $\mathit{\{R_5,C_1,C_2\}}$, we take the output of $\mathit{R_5}$ to be feature representations in experiments. The detailed procedure of an HCN-based person re-ID is shown in Figure~\ref{flowchart}.

\label{ssec:hcn}
\subsection{Combined dataset}
\label{ssec:joint}
According to the setting of person re-ID, there are differences between training and probe images. Training images belong to one of the trained classes but probe images do not belong to any existing classes since they are usually "unseen", i.e., $\mathit{d_x\in{M}, d_p\in{K}}$, $M\cap{K}=\emptyset$, where $\mathit{d_x}$ is the identity of training image $\mathit{x}$.

As the above-mentioned situation, training multi-datasets jointly is feasible. Since the training dataset is disjoint with the probe images. Suppose we have $\mathit{D}$ datasets, each dataset has $\mathit{M_i}$ identities and $\mathit{N_i}$ images. All training images are denoted as $\mathit{\{(x_{ij},d_{ij})_{j=1}^{N_i}\}_{i=1}^{D}}$, where $\mathit{x_{ij}}$ means the $\mathit{i}$-th image in the $\mathit{j}$-th dataset and $\mathit{d_{ij}}$ denotes the identity of $\mathit{x_{ij}}$ where $\mathit{d_{ij}\in{\{1,2,...,M_i\}}}$.
 To avoid potential conflict on an id number, datasets are merged together into a combined dataset $\mathit{N = \sum_{i=1}^{D}N_i}$ with new label $\mathit{d}\in{\{1,2,...,M\}}$. This combined dataset which has $\mathit{M}$ identities would be used in our framework during the training stage.
\subsection{Re-ranking}
\label{ssec:re_rank}
%This work aims to receive higher accuracy by an adding step which makes ranking list to consist more relevant images. There are many methods could be applied on person re-ID\cite{qin2011hello,zheng2015query,ye2016person,zhong2017re}, we follow the state-of-the-art method\cite{zhong2017re}.

Given a probe $\mathit{p}$ and a gallery set with $\mathit{K}$ images $\mathit{G=\{g_i\}_{i=1}^K}$, the initial ranking list $\mathit{L(p,G) = \{g_1,g_2,...g_K\}}$ is obtained according to the pairwise distance $\mathit{D_{p,g_i} =\sqrt{ (f_p-f_{gi})^2}}$ where $\mathit{f_p}$ and $\mathit{f_{gi}}$ are computed feature representations of probe $\mathit{p}$ and gallery $\mathit{g_i}$, respectively. Following \cite{qin2011hello}, the $\mathit{k}$-reciprocal nearest neighbors $\mathit{R(p,k)}$ can be defined as,
\begin{equation}
\mathit{R(p,k)=\{g_i|(g_i\in{N(p,k)})\wedge(p\in{N(g_i,k)})\}}
\end{equation}
,where $\mathit{N(p,k)}$ is the $\mathit{k}$-nearest neighbors of a probe $\mathit{p}$,
\begin{equation}
\mathit{N(p,k) = \{n_1,n_2,...n_k\}, N(p,k)\subset{G}, |N(p,k)|=k}
\end{equation}
To increase relevant images which are more similar than $\mathit{R(p,k)}$ to probe $\mathit{p}$, we follow \cite{zhong2017re} to add additional $\frac{1}{2}\mathit{k}$-reciprocal nearest neighbors to each $\mathit{R(p,k)}$ to form $\mathit{R^*(p,k)}$ and re-calculate the pairwise distance between the probe $\mathit{p}$ and gallery $\mathit{g_i}$ by combining both Euclidean and Jacaard distance~\cite{jaccard1912distribution},
\begin{equation}
d(p,g_i)=(1-\lambda)d_J(p,g_i)+\lambda d(p,g_i)
\end{equation}
where the Jaccard distance of the k-reciprocal set is defined as
\begin{equation}
\mathit{d_J(p,g_i)=1-\frac{|R^*(p,k)\cap{R^*(g_i,k)|}}{|R^*(p,k)\cup{R^*(g_i,k)}|}}
\end{equation}
And$|\cdot|$ denotes the cardinality of the set. Methods mentioned above make ranking lists contain more hard positive images that can be used to increase performance.

\section{experiments}
\label{sec:typestyle}
%%%%%%%%%%%%%%%%%%%%%%%%%%%%fig
\begin{figure*}[tbp]
 
\begin{minipage}[t]{.3\linewidth}
  \centering
  \centerline{\includegraphics[width=6.0cm]{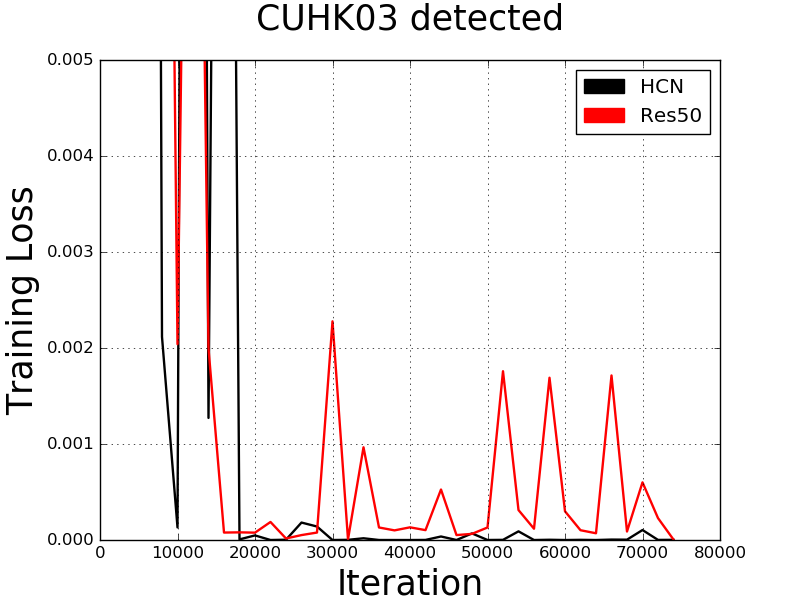}}
%  \vspace{2.0cm}
  \centerline{(a)}\medskip
\end{minipage}
\hfill
\begin{minipage}[t]{0.3\linewidth}
  \centering
  \centerline{\includegraphics[width=6.0cm]{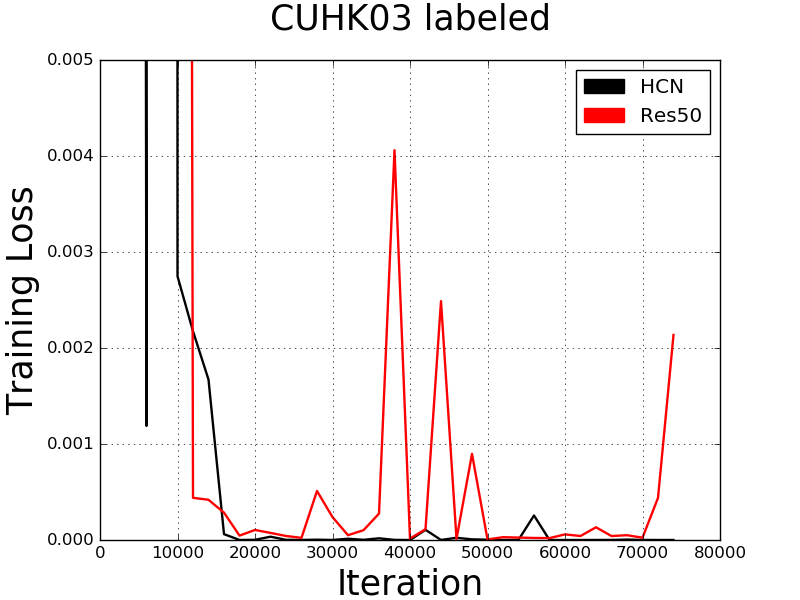}}
%  \vspace{1.5cm}
  \centerline{(b)}\medskip
\end{minipage}
\hfill
\begin{minipage}[t]{0.3\linewidth}
  \centering
  \centerline{\includegraphics[width=6.0cm]{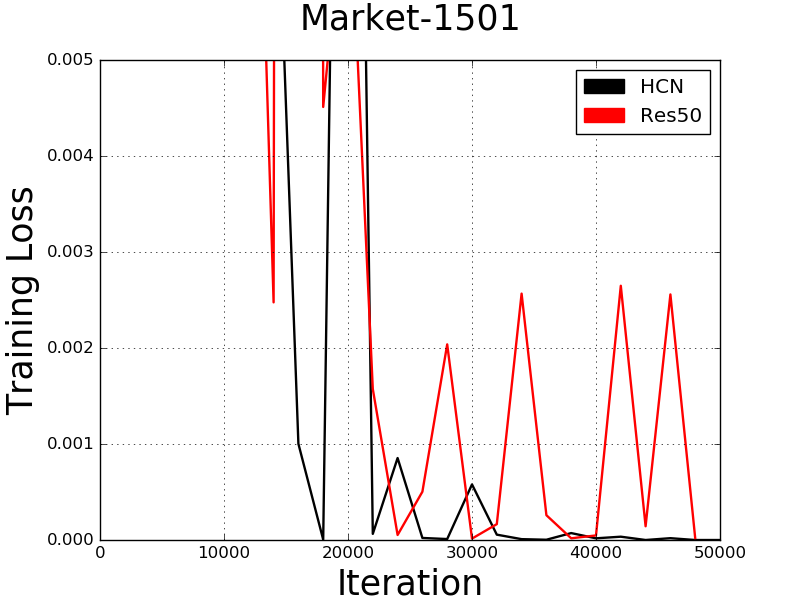}}
%  \vspace{1.5cm}
  \centerline{(c)}\medskip
\end{minipage}
 
\caption{Comparisons on training loss of $\mathit{R_5}$ between HCN and ResNet-50(backbone) on datasets: (a) CUHK03 detected dataset. (b) CUHK03 labeled dataset. (c) Market-1501 dataset. }
\label{HCN_loss}
\end{figure*}
%%%%%%%%%%%%%%%%%%%%%%%%%%%%%%%%%%%%%% table
\begin{table}[tbp]
\centering
\begin{tabular}{|c|c|c|c|c|c|c|c|c|}
\hline
\multicolumn{2}{|l|}{Datasets} & \multicolumn{2}{|c|}{\# ID}
& \multicolumn{2}{|c|}{\# Training ID} & \multicolumn{2}{|c|}{\# Testing ID}\\
\hline\hline
%previous
\multicolumn{2}{|l|}{Market-1501} & \multicolumn{2}{|c|}{1501}
& \multicolumn{2}{|c|}{751} & \multicolumn{2}{|c|}{750}\\
\multicolumn{2}{|l|}{CUHK03} & \multicolumn{2}{|c|}{1467}
& \multicolumn{2}{|c|}{767} & \multicolumn{2}{|c|}{700}\\
\multicolumn{2}{|l|}{CUHK01} & \multicolumn{2}{|c|}{971}
& \multicolumn{2}{|c|}{971} & \multicolumn{2}{|c|}{NaN}\\
\hline
\multicolumn{2}{|l|}{Combined} & \multicolumn{2}{|c|}{2489}
& \multicolumn{2}{|c|}{2489}
& \multicolumn{2}{|c|}{NaN}\\
\hline
\end{tabular}
\caption{The details of number of identities on datasets.}
\label{datasets}
\end{table}
%%%%%%%%%%%%%%%%%%%%%%%%%%%%%%%%%%%%%
\subsection{Datasets and settings}
\label{ssec:datasets}
Two large image-based datasets are adopted in experiments for closing to the realistic settings. Market-1501~\cite{MARKET} is currently the largest image-based dataset which is prepared by a university. It has 32,668 images which include 1501 identities that are detected by Deformable Part Model (DPM)~\cite{DPM}. The dataset is divided into two non-overlapping parts, 751 identities with 12,936 images and 750 identities with 19,732 images for training and testing respectively. The first set of experiments are conducted based on single-query inputs.

CUHK03~\cite{CUHK03} is also a large scale image-based dataset which is formed by 14,096 images with 1467 identities. Among there identities, 767 are used for training and 700 are used for testing. Following the protocol~\cite{zhong2017re} , CUHK03 is split into two subsets which are CUHK03 labeled and CUHK03 detected based on the methods adopted for detecting person in images. Specifically, CUHK03 is manually labeled and CUHK03 detected is the formed by applying DPM. CUHK01~\cite{CUHK01} is also an image-based dataset and its images are also captured in the same university. It is composed of 971 identities and we add all 971 identities to the combined dataset in the training stage for augementing data. Details is shown in Table~\ref{datasets}.

\subsection{Evaluation protocal}
\label{ssec:evaluation}
\begin{figure}[tbp]

\begin{minipage}[b]{1.0\linewidth}
  \centering
  \centerline{\includegraphics[width=8.0cm]{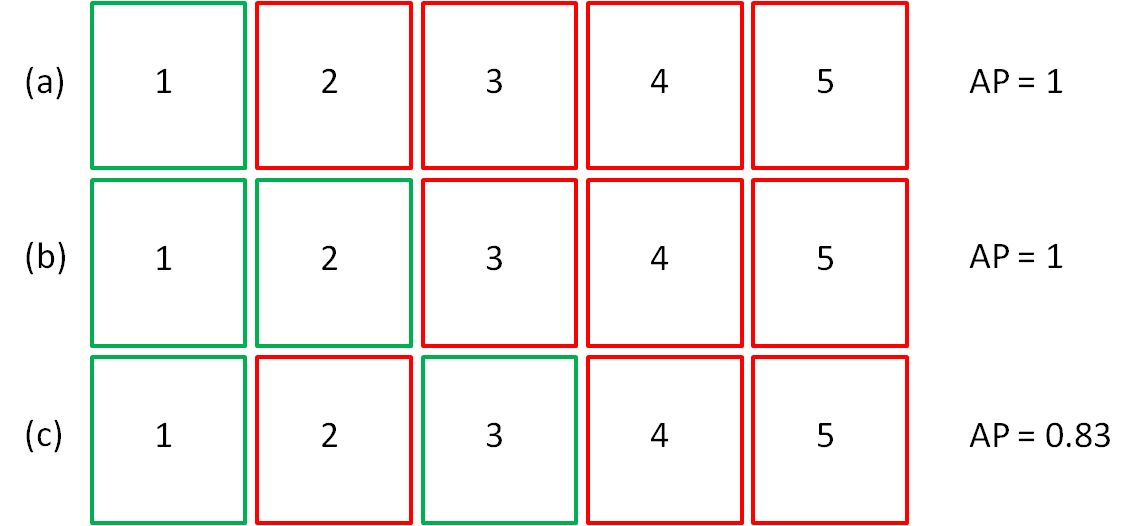}}
  \vspace{0.1cm}
  \centerline\medskip
\end{minipage}
\caption{An example of difference between AP and CMC. True matches are in green boxes and red boxes denote false matches. For rank list (a) (b) and (c), the evaluation of CMC are all 1. But AP will return the value of 1, 1, and 0.83, respectively}
\label{example_mAP}
\end{figure}

Many datasets use Cumulated Matching Characteristics (CMC) curve to evaluate methods of person re-ID. A CMC curve is used to measure accuracy performance of methods that produce the rank list of possible matches and they usually turn out with the format of rankings. If there is only ground truth image for a given query image which is shown in Figure~\ref{example_mAP} (a), evaluations usually focus only on the precision rate, and under the circumstance the CMC curve is effictive. However, if there are more than one ground truth images in response to a given query image shown in Figure~\ref{example_mAP} (b) and (c), then mean Average Percision (mAP) is more appropriate for providing fair comparisons. In this case, both precision and recall have to be considered, and the precision of good methods will stay high as the recall rate increases. In this work, both Rank-1 and mAP are applied to evaluate the proposed methods.
%%%%%%%%%%%%%%%%%%%%%%%%%%%%%%%%%%%%%% table 
\begin{table}[tbp]
\centering
\begin{tabular}{|c|c|c|c|c|c|c|}
\hline
\multicolumn{3}{|c|}{\multirow{2}{*}{Methods}}&
\multicolumn{2}{c|}{CUHK03 detected} &
\multicolumn{2}{c|}{CUHK03 labeled} \\
\cline{4-7}
\multicolumn{3}{|c|}{} & Rank-1 & mAP & Rank-1 & mAP \\
\hline\hline
%previous
\multicolumn{3}{|l|}{LOMO+XQDA~\cite{liao2015person}} & 16.6 & 17.8 & 19.1 & 20.8 \\
\multicolumn{3}{|l|}{XQDA~\cite{zhong2017re}} & 31.1 & 28.2 & 32.0 & 29.6 \\
\multicolumn{3}{|l|}{XQDA+re-rank~\cite{zhong2017re}} & 34.7 & 37.4 & 38.1 & 40.3 \\
\hline
%HCN
%\multicolumn{3}{|l|}{HCN} & 34.9 & 32.1 & 35.6 & 34.1 \\
\multicolumn{3}{|l|}{HCN+re-rank} & 40.9 & 43.0 & 43.4 & 46.0 \\
\hline
%ours
%\multicolumn{3}{|l|}{Ours+XQDA} & 36.0 & 32.2 & 38.3 & 34.1 \\
\multicolumn{3}{|l|}{Ours+XQDA+re-rank} & \textbf{43.7} & \textbf{45.3} & \textbf{44.0} & \textbf{46.9} \\
\hline
\end{tabular}
\caption{Comparison of previous methods with our results on CUHK03 detected and CUHK03 labeled datasets.}
\label{CUHK03}
\end{table}
%%%%%%%%%%%%%%%%%%%%%%%%%%%%%%%%%%%%%%% table 
\begin{table}[tbp]
\centering
\begin{tabular}{|c|c|c|c|c|c|c|}
\hline
\multicolumn{3}{|c|}{\multirow{2}{*}{Methods}}&
\multicolumn{4}{c|}{Market-1501}\\
\cline{4-7}
\multicolumn{3}{|c|}{} & 
\multicolumn{2}{|c|}{Rank-1} & 
\multicolumn{2}{|c|}{mAP}\\
\hline\hline
%previous
\multicolumn{3}{|l|}{SCSP~\cite{chen2016similarity}} & \multicolumn{2}{|c|}{51.90} & \multicolumn{2}{|c|}{26.35} \\
\multicolumn{3}{|l|}{DNS~\cite{zhang2016learning}} & \multicolumn{2}{|c|}{61.02} & \multicolumn{2}{|c|}{35.68} \\
\multicolumn{3}{|l|}{Gated~\cite{varior2016gated}} & \multicolumn{2}{|c|}{65.88} & \multicolumn{2}{|c|}{39.55} \\
\multicolumn{3}{|l|}{ResNet50+Euclidean+re-rank\cite{zhong2017re}} & \multicolumn{2}{|c|}{81.44} & \multicolumn{2}{|c|}{70.39} \\
%ours
\hline
\multicolumn{3}{|l|}{HCN+re-rank} & \multicolumn{2}{|c|}{84.09} & \multicolumn{2}{|c|}{74.28} \\
\hline
\multicolumn{3}{|l|}{Ours+re-rank} & \multicolumn{2}{|c|}{\textbf{82.63}} & \multicolumn{2}{|c|}{\textbf{72.66}} \\
\hline
\end{tabular}
\caption{Comparison of previous methods with our results on Market-1501 dataset.}
\label{Market}
\end{table}
%%%%%%%%%%%%%%%%%%%%%%%%%%%%%%%%%%%%%%%

\subsection{Comparison with state-of-the-art methods}
\label{ssec:stateoftheart}

We first evaluate the performance of HCN and show the results in Table \ref{CUHK03} and Table \ref{Market}. HCN obtains better performance on Rank-1 and mAP than \cite{zhong2017re} on CUHK03 detected, CUHK03 labeled and Market-1501 datasets. To explore the effect of hierarchical feature maps on HCN, we compare the training loss of the prediction of $\mathit{R_5}$ between HCN and ResNet-50 in Figure \ref{HCN_loss}. Under the same iteration, HCN has lower training loss than ResNet-50. This outcome indicates the hierarchical feature maps $\mathit{\{C_1, C_2\}}$ really help train better feature representations and also speed up the convergence rate.

We also compare our method (training HCN and combined dataset jointly) with state-of-the art methods on CUHK03 labeled and CUHK03 detected datasets and show the results in Table \ref{CUHK03}. XQDA metric is shown since methods~\cite{zhong2017re,liao2015person} have the best rank-1 and mAP for this metric. In the CUHK03 detected dataset, our method gets 43.7\% and 45.3\% accuracy on rank-1 and mAP respectively, which gains more than 9\% and 7.9\% improvement compared with~\cite{zhong2017re,liao2015person} methods. In the CUHK03 labeled dataset, our method obtains better results than previous methods, we have 44.0\% gain for rank-1 and 46.9\% gain for mAP.

When the experiments are conducted based on the Market-1501 dataset, the results shown in Table~\ref{Market} are those obtained by our method and other state-of-the-art methods. It is clear that our method outperforms the methods proposed by~\cite{chen2016similarity},~\cite{zhang2016learning},~\cite{varior2016gated} and~\cite{zhong2017re}. Among the four state-of-the-art methods, the method that uses the combination of ResNet50+ Euclidean+ Re-rank~\cite{zhong2017re} receives the best Rank-1 (81.44\%) and mAP (70.39\%). However, our method beats it in both the Rank-1 (82.63\%) and mAP (72.66\%) results. As to the effect of combined dataset, originally we expect to get better performance due to the potential of having more general feature representations. Unfortunately the received rank-1 and mAP results are both slightly lower than our HCN+ re-rank method we proposed in the Market-1501 dataset case. This result meets the hypothesis made in~\cite{dgd}. In~\cite{dgd}, it was reported that a large domain helps a smaller domain dataset on learning, but would degrade the retrieval capability at the same time.

%%%%%%%%%%%%%%%%%%%%%%%%%%%%%%%%%%%%%%%%% table 
\begin{table}[tbp]
\centering
\begin{tabular}{|c|c|c|c|c|c|c|c|c|c|c|}
\hline
\multicolumn{3}{|c|}{Methods}&
\multicolumn{2}{c|}{detected} &
\multicolumn{2}{c|}{labeled} &
\multicolumn{2}{c|}{Market-1501} &
\multicolumn{2}{c|}{Total} \\
\hline\hline
%previous
\multicolumn{3}{|l|}{Zhong \cite{zhong2017re}} & \multicolumn{2}{|c|}{36} & \multicolumn{2}{|c|}{36} & \multicolumn{2}{|c|}{24} & \multicolumn{2}{|c|}{96} \\
%ours
\hline
\multicolumn{3}{|l|}{Our} & \multicolumn{6}{|c|}{{43}} & \multicolumn{2}{|c|}{\textbf{43}} \\
\hline
\end{tabular}
\caption{Training time (hour) of Zhong [13] and our method on CUHK03 detected, CUHK03 labeled and Market-1501.}
\label{time}
\end{table}

The advantage of using a combined dataset for training is to save the training time. In the experiments, the training time consumed on training different datasets are illustrated in Table \ref{time}. We use a Titan X GPU to run settings mentioned above and count the period between the start of training and the end of training according to its own log files. Costs on CUHK03 detected, CUHK03 labeled and Market-1501 datasets are about 36 hours, 36 hours and 24 hours respectively. Altogether, it costs almost 96 hours for finishing all trains. In our work, the combined dataset needs about 43 hours which is longer than individual datasets. However, to satisfy requirements in experiments, only one model is needed due to generalized features could be applied on cross datasets. Comparing with 96 hours, it saves more than 50\% computation time.

\begin{figure}[tbp]

\begin{minipage}[b]{1.0\linewidth}
  \centering
  \centerline{\includegraphics[width=8.5cm]{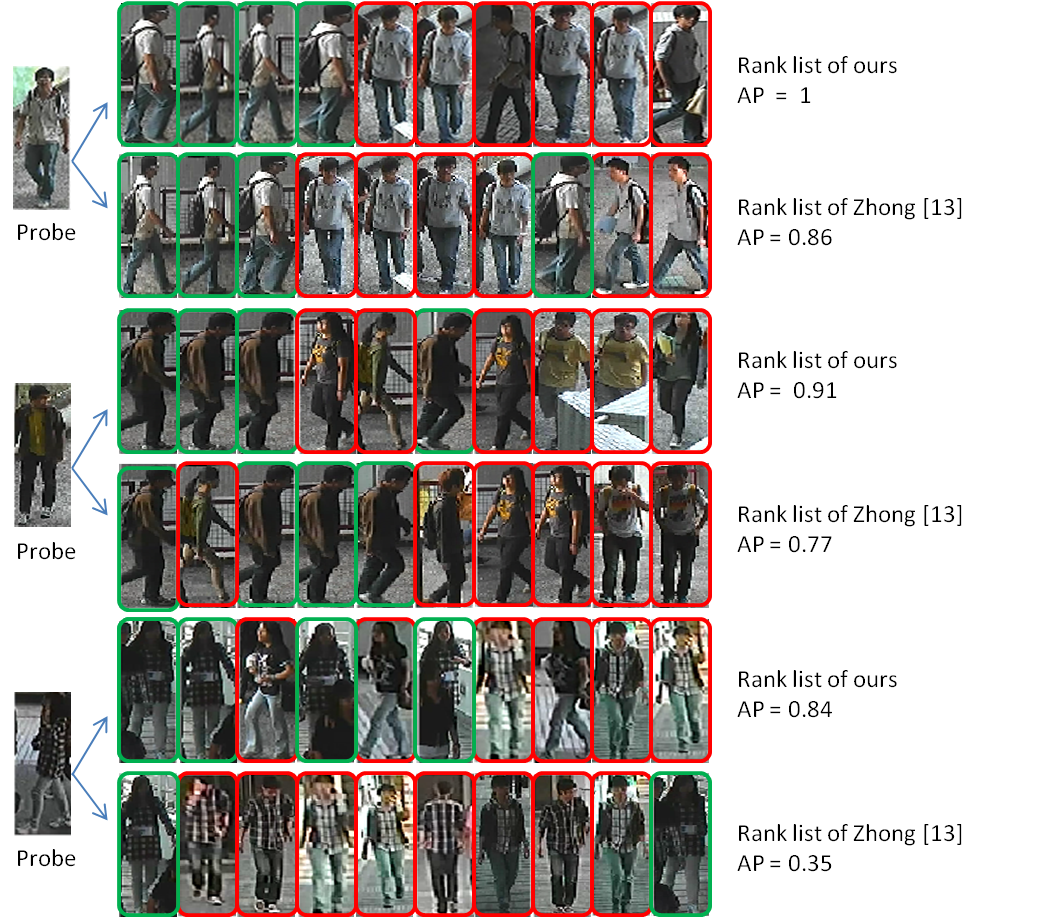}}
  \vspace{0.1cm}
  \centerline\medskip
\end{minipage}
\caption{Experiment results on CUHK03 labeled dataset. Images surrounded by green and red rectangles denote correct and incorrect matches, respectively.}
\label{comparision}
\end{figure}

In Figure~\ref{comparision}, some experiment results on CUHK03 labeled are shown. Our method and the method of Zhong~\cite{zhong2017re} are shown on the upper and the lower lists, respectively, for three different probe images.

%%%%%%%%%%%%%%%%%%%%%%%%%%%%fig
%%%%%%%%%%%%%%%%%%%%%%%%%%%

%%%%%%%%%%%%%%%%%%%%%%%%%%%%fig
%%%%%%%%%%%%%%%%%%%%%%%%%%%

\section{conclusions}
\label{sec:majhead}
Person re-ID is a challenging problem. To learn better feature representations, we applied HCN for learning high semantic features and combined dataset for making features being more general. Experiments show our method outperformed state-of-the-art methods. For future work, handling video-based person re-ID or person re-ID in a whole scene image will be interesting applications for us.

% Below is an example of how to insert images. Delete the ``\vspace'' line,
% uncomment the preceding line ``\centerline...'' and replace ``imageX.ps''
% with a suitable PostScript file name.
% ------------------------------------------------------------------------

%\vfill
%\pagebreak

%\hfill\vfill\pagebreak

%\section{REFERENCES}
%\label{sec:refs}


\end{document}